\documentclass{article}
\usepackage[margin=1in]{geometry}
\usepackage[utf8]{inputenc}
\usepackage{graphicx}
\usepackage{hyperref}
\usepackage{caption}
\usepackage{subcaption}
\usepackage{booktabs}
\usepackage{tabularx}
\usepackage{amsmath}
\usepackage{wrapfig}
\usepackage{color}
\usepackage{svg}
\usepackage{wrapfig}
\usepackage{enumitem}
\usepackage{authblk}

\title{An evaluation of deep learning models for predicting water depth evolution in urban floods}

\author[2]{Stefania Russo}
\author[1]{Nathana\"el Perraudin}
\author[1]{Steven Stalder}
\author[1]{Fernando Perez-Cruz}
\author[3]{Joao Paulo Leitao}
\author[1]{Guillaume Obozinski}
\author[4]{Jan Dirk Wegner}
\affil[1]{Swiss Data Science Center, ETH Z\"urich and EPF Lausanne, Switzerland}
\affil[2]{Photogrammetry and Remote Sensing group, ETH Zurich, Switzerland}
\affil[3]{Department of Urban Water Management, EAWAG, Switzerland}
\affil[4]{Institute for Computational Science, University of Zurich, Switzerland}

\date{August 2022}

\begin{document}

\maketitle

\section*{Abstract}
In this technical report we compare different deep learning models for prediction of water depth rasters at high spatial resolution.
Efficient, accurate, and fast methods for water depth prediction are nowadays important as urban floods are  increasing due to higher rainfall intensity caused by climate change, expansion of cities and changes in land use. 
While hydrodynamic models models can provide reliable forecasts by simulating water depth at every location of a catchment, they also have a high computational burden which jeopardizes their application to real-time prediction in large urban areas at high spatial resolution.
Here, we propose to address this issue by using data-driven techniques. Specifically, we evaluate deep learning models which are trained to reproduce the data simulated by the CADDIES cellular-automata flood model, providing flood forecasts that can occur at different future time horizons. 
The advantage of using such models is that they can learn the underlying physical phenomena \textit{a priori}, preventing manual parameter setting and computational burden.
We perform experiments on a dataset consisting of two catchments areas within Switzerland with 18 simpler, short rainfall patterns and 4 long, more complex ones.  
Our results show that the deep learning models present in general lower errors compared to the other methods, especially for water depths $>0.5m$. However, when testing on more complex rainfall events or unseen catchment areas, the deep models do not show benefits over the simpler ones.


\section{Introduction}

Floods are one of the most common natural disasters with an unpredictable and destructive force carrying devastating consequences for infrastructure, economy and societies \cite{int_35}. At the whim of nature, floods can affect massive areas leaving no possibility to prevent or undo the destruction following the disaster. 
In urban areas, floods occur when natural water sources or drainage systems in cities lack the capacity to convey excess water caused by intense rainfall events~\cite{int_28}. 
Thus, early warning systems through real-time flood predictions~\cite{int_33}, and a quick and effective response from rescue services make up the most essential counter-measures to floods~\cite{int_2}.
While hydrodynamic flood simulation models can provide reliable forecasts for water depth at every location of a catchment, their high computational burden, as well as required expertise and in-depth knowledge about hydrological parameters~\cite{rwf_1} is hindering their application.
For this reason, data-driven models have been developed in recent years for flood forecasting and prediction \cite{prev_1, int_26, int_3, int_5},
In this technical report, we compare several deep learning models for dense water depth prediction at different future time horizons. The models only require terrain features such as the Digital Elevation Model (DEM) of the area under study and a rainfall forecast.

We evaluate the models on a dataset consisting of two catchments areas within Switzerland with 18 simple, short rainfall patterns and 4 long, more complex ones.
Our results show that the deep learning models present in general lower errors compared to the other methods, especially for water depths $>0.5m$. 
On the other hand, the deep models do not show any benefits over the simpler one when testing on more complex rainfall events or unseen catchment areas. 
We discuss that this is probably caused by the fact that the networks only capture the easiest correlations in the dataset so and fail to do so for the complex dynamics in the long rainfall events. 
We conclude this report proposing potential solutions for future works.

\section{Related Works}
Flooding is and has been a massive concern over past decades, causing wide-spread damage to infrastructure and economy. Greatest damages can arise especially in the case of urban flash-floods where the reaction time and planning time is limited. Predicting large scale flash-floods is therefore crucial to support counter measures such as early warning systems.
In this section, an overview of techniques for flood prediction is given.

Hydrodynamic models have been in use for long time to predict water-related events, such as floods~\cite{rwf_10}, storms~\cite{rwf_2, rwf_3}, rainfall/runoff~\cite{rwf_4, rwf_5} and flows~\cite{rwf_7, rwf_8}.
Physically-based hydrodynamic models have been used for a long time in urban flood simulation~\cite{bradbrook2004two, chen2007urban, bates2010simple}. Nevertheless, the development of physically-based models often requires expertise related to hydrological parameters, making the application to spatial flood forecasting challenging~\cite{kim2015urban}, in addition to its high computational cost. Other approaches used for flood prediction are simple and easy-to-use statistical models, such as multiple criteria decision methods, weights-of-evidence and flood frequency analysis~\cite{panahi2021deep}. However, due to various simplifying assumptions, they often produce results with insufficient precision and tend to be unreliable.
Another group of models for flood estimation are the so-called physically-simplified approaches \cite{rwf_17, rwf_18} predicting floods through simplified hydraulic notions. As a result, these model provide predictions with much lower computational costs~\cite{rwf_19, rwf_20}. The cellular-automata flood models \cite{dat_1} have recently gained attention by the hydrological community. Instead of solving complex shallow water equations, these models carry out faster flood modelling by using transition rules which work by predicting the new state (i.e. the amount of water in each cell) based on the cell's previous state and its neighbors. This is applied on all raster cells in parallel, thus supporting GPU parallelism, greatly reducing the simulation time~\cite{dat_1} at the cost of accuracy. 

The shortcomings of the above mentioned hydrodynamic models encouraged the use of advanced data-driven models, such as Machine Learning (ML). A further reason for the popularity of ML is that it can numerically formulate the nonlinear characteristics of the flood without the need to learn the physical processes from scratch~\cite{mosavi2018flood}. Many ML models, such as multilayer perceptron~\cite{rezaeian2010daily}, wavelet neural networks~\cite{nourani2014applications}, support vector machine~\cite{tehrany2015flood, choubin2019ensemble}, decision tree based classifiers~\cite{khosravi2018comparative, janizadeh2019prediction}, neuro-fuzzy~\cite{mosavi2018hybrid}, and artificial neural networks (ANNs)~\cite{janizadeh2019prediction}, have been applied to flood forecasting.

In recent years, Deep Learning (DL) models, which fall into the general umbrella of ML, have gained more popularity since they overcome many of the limitations of traditional ML methods on the modelling process~\cite{van2020spatially}. Compared to earlier ANNs, DL models can identify complex nonlinear relationships underlying predictor and outcome variables~\cite{panahi2021deep}, which also enables DL models to outperform traditional ML methods in flood forecasting tasks successfully. 
In~\cite{rwf_22}, the authors forecast a river stream flow based on the rainfall measurements from several gauge stations using a recurrent neural network. This method was later extended to multiple-step-ahead using an expandable ANN architecture in~\cite{rwf_21}. In~\cite{rwf_23}, a single long short-term memory model was trained on hundreds of basins using meteorological time series data and static catchment attributes. Their approach significantly outperformed hydrological models that were calibrated regionally. 
\cite{prev_1} used a convolutional neural network (CNN) model to perform a daily runoff prediction. Similarly, \cite{int_26} employed a CNN to predict the maximum water depth maps (i.e. an hazard map) in urban pluvial flood events. 
\cite{int_3} also used a CNN-based U-Net~\cite{ronneberger2015u} model, exploring the where the generalization capabilities of CNNs as flood prediction models. 
Recently, in \cite{int_5}, DL techniques were used for predicting gauge height, with higher accuracy than the physical and statistical models currently in use. \\
While most of the above works focus on predicting the maximum water depth that can occur for any point in the map in the near future, in this technical report we consider the task of providing flood forecasts at different future time horizons (i.e. 30\,minutes, 1\,hour). 
In this way, we aim at modeling the full evolution of water depth at different time steps.

\section{Experiment setup}
\setlength{\intextsep}{0pt}%
\begin{figure}[ht!]
    \centering
    \includegraphics[width=0.3\columnwidth]{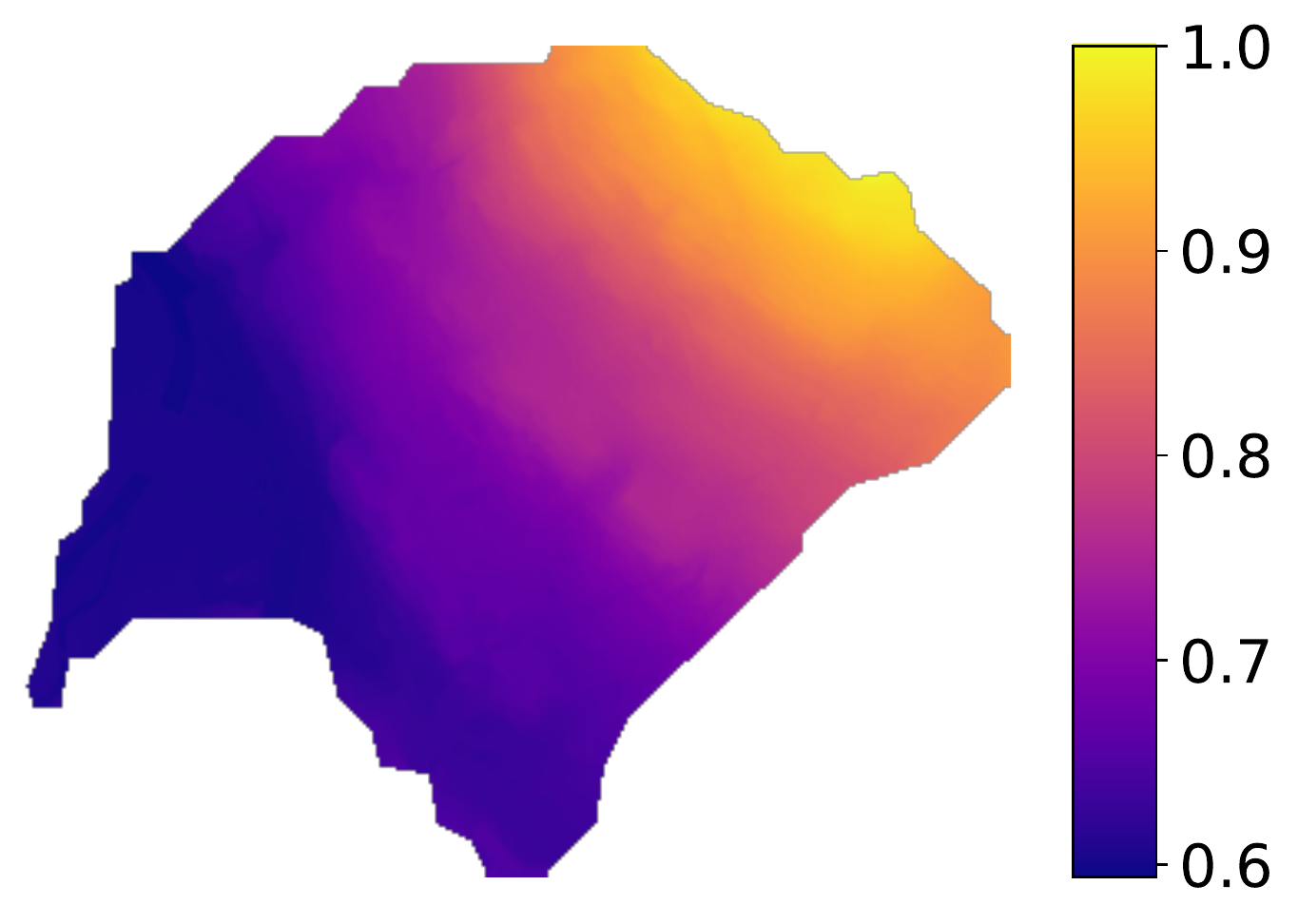}
    \caption{DEM (normalized elevation) of the main catchment area 709. Dark to light color indicates low to high elevation. The values that are completely zero are not part of the catchment area.}
    \label{fig:dem}
\end{figure}
\setlength{\intextsep}{12pt}%

\subsection{Dataset}
For our experiments, we use a dataset consisting of simulations run by the CADDIES cellular-automata flood model~\cite{guidolin2016weighted}. 
The transformation rules of the CADDIES model establish flow motions through a weight-based mechanism, which avoid the need to solve complicated and time-consuming equations. It outperforms physically-based models for solving shallow water equations in terms of computational performance, with a decrease in accuracy~\cite{guidolin2016weighted}. 
Using this dataset allows us to conveniently verify our approach with low computational effort. The simulated results consist of pairings of rainfall events and water depths in two catchment areas across Switzerland, which are named 709 and 744 according to a DEM numbering system. The data comes with a spatial resolution of 1m.

\begin{figure}[t!]
     \centering
     \begin{subfigure}[b]{0.3\columnwidth}
         \centering
         \includegraphics[width=\textwidth]{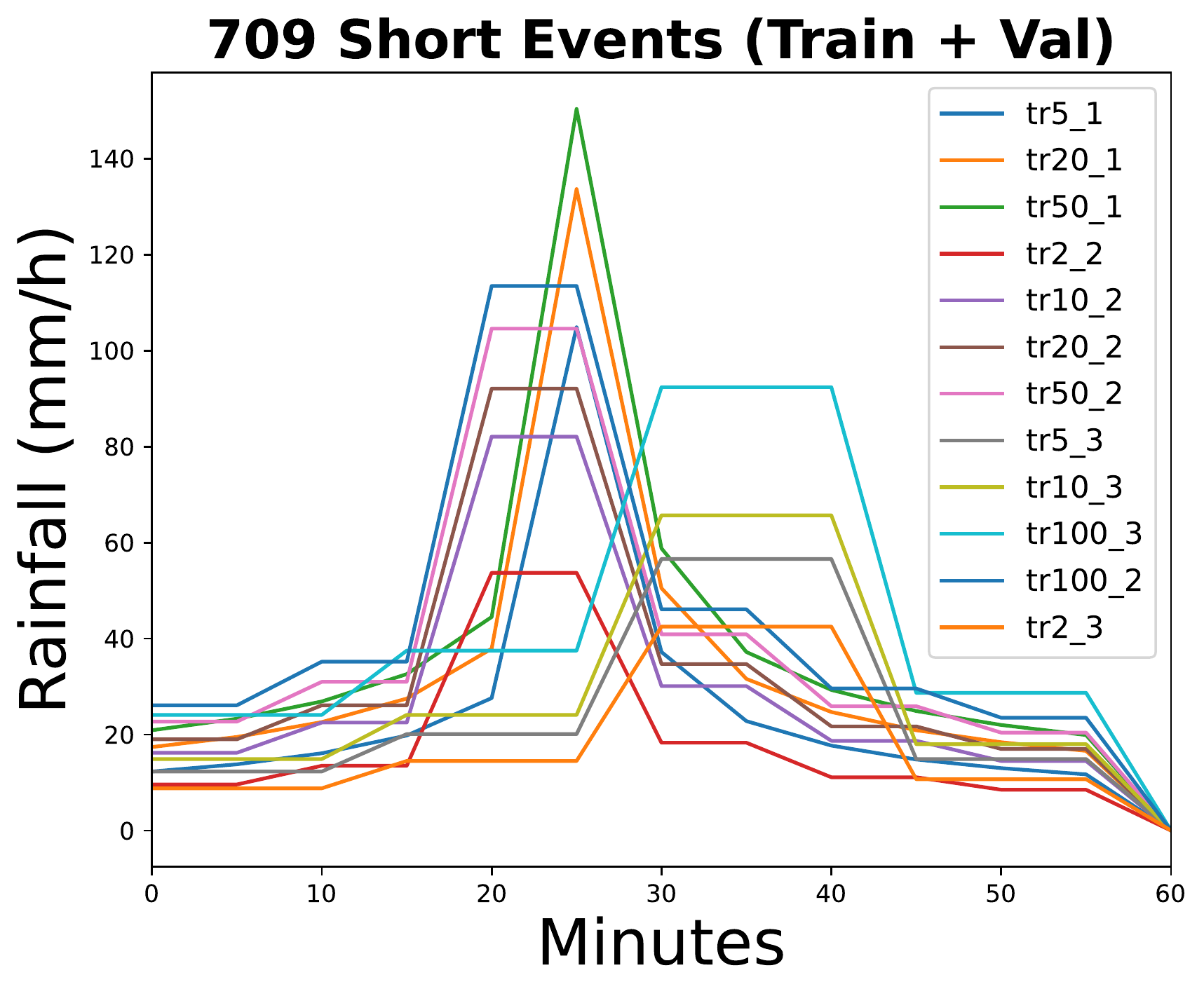}
     \end{subfigure}
     \hfill
     \begin{subfigure}[b]{0.3\columnwidth}
         \centering
         \includegraphics[width=\textwidth]{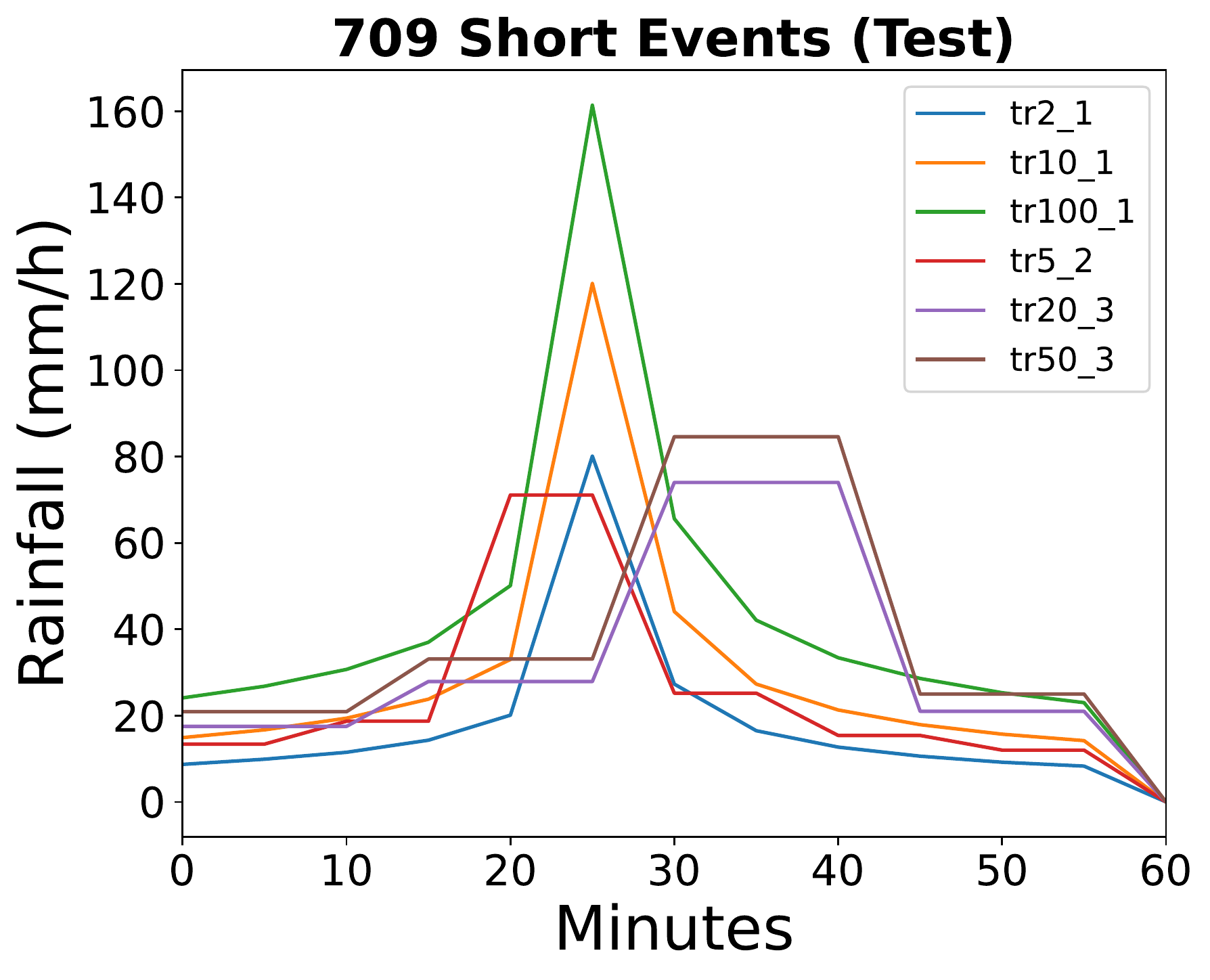}
     \end{subfigure}
     \hfill
     \begin{subfigure}[b]{0.3\columnwidth}
         \centering
         \includegraphics[width=\textwidth]{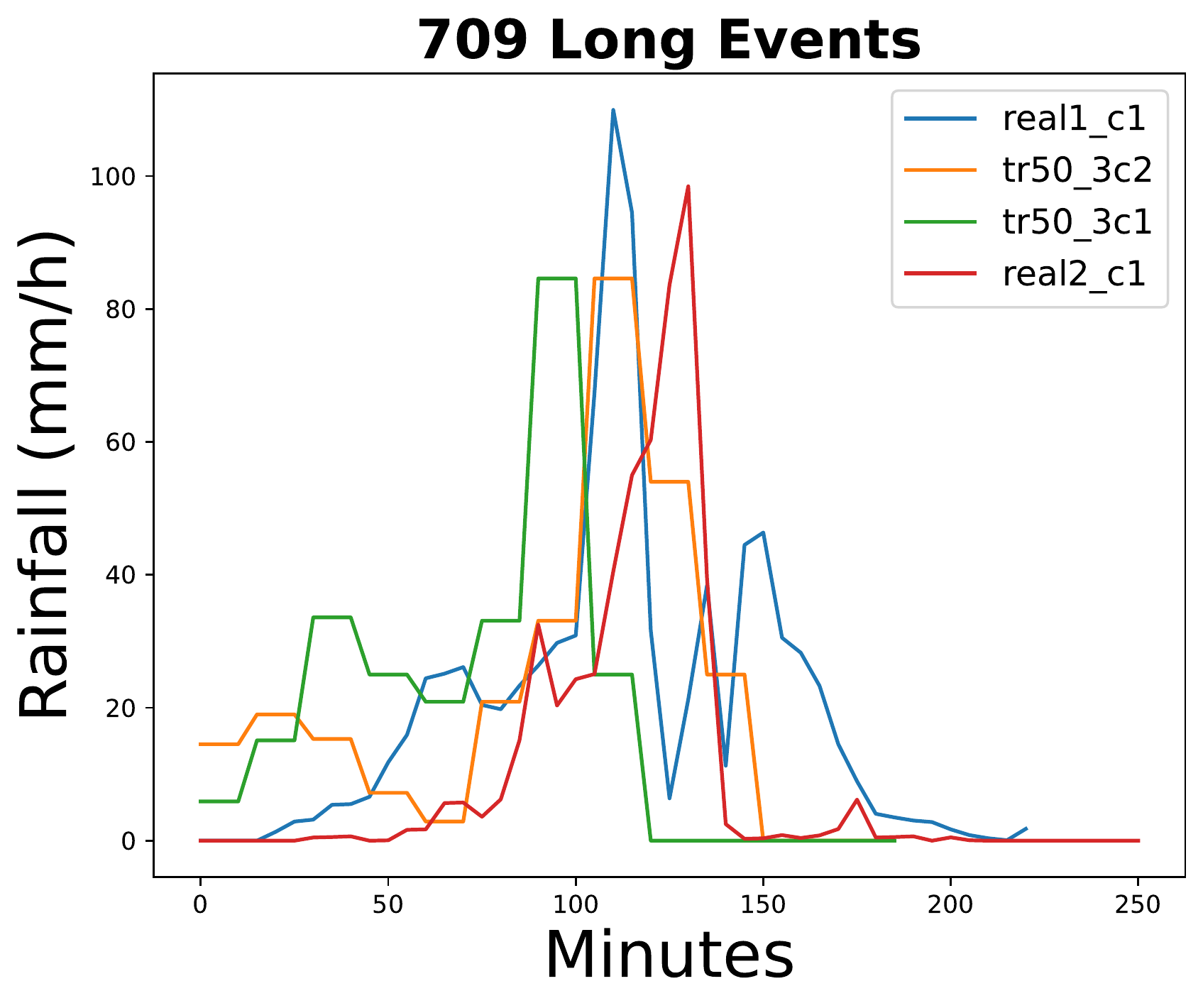}
     \end{subfigure}
     \caption{Hyetographs of rainfall events used for simulations. For the short events, the first number of the label denotes the return period, while the second denotes the time step discretization (5 minutes - "1", 10 minutes - "2" and 15 minutes - "3"). For the long events: the same naming is applied, in addition to "c" that stands for "continued" rainfall. Additionally, two real long rainfall events were also used, \textit{real1\_c1} and \textit{real2\_c1}, lasting two hours and with a return period of approximately 20 years.}
     \label{fig:hyetographs}
\end{figure}

\paragraph{Data Preparation}\label{sec:pre}
Most experiments are performed on the catchment area 709 from the dataset, whose elevation map we show in Fig.~\ref{fig:dem}. Eighteen different one-hour short rainfall events in this catchment area are created based on the return periods of 2, 5, 10, 20, 50, and 100 years. Additionally, we have access to four long rainfall events over time periods of up to four hours. In Figure~\ref{fig:hyetographs}, we show the amount of rainfall for the short events, split into training and test sets, as well as for all long events, which we individually add to the training or test set for some experiments.
The simulation are discretized with a time step of 5 minutes.

\begin{wrapfigure}{r}{0.25\textwidth}
    \centering
    \includegraphics[width=0.25\columnwidth]{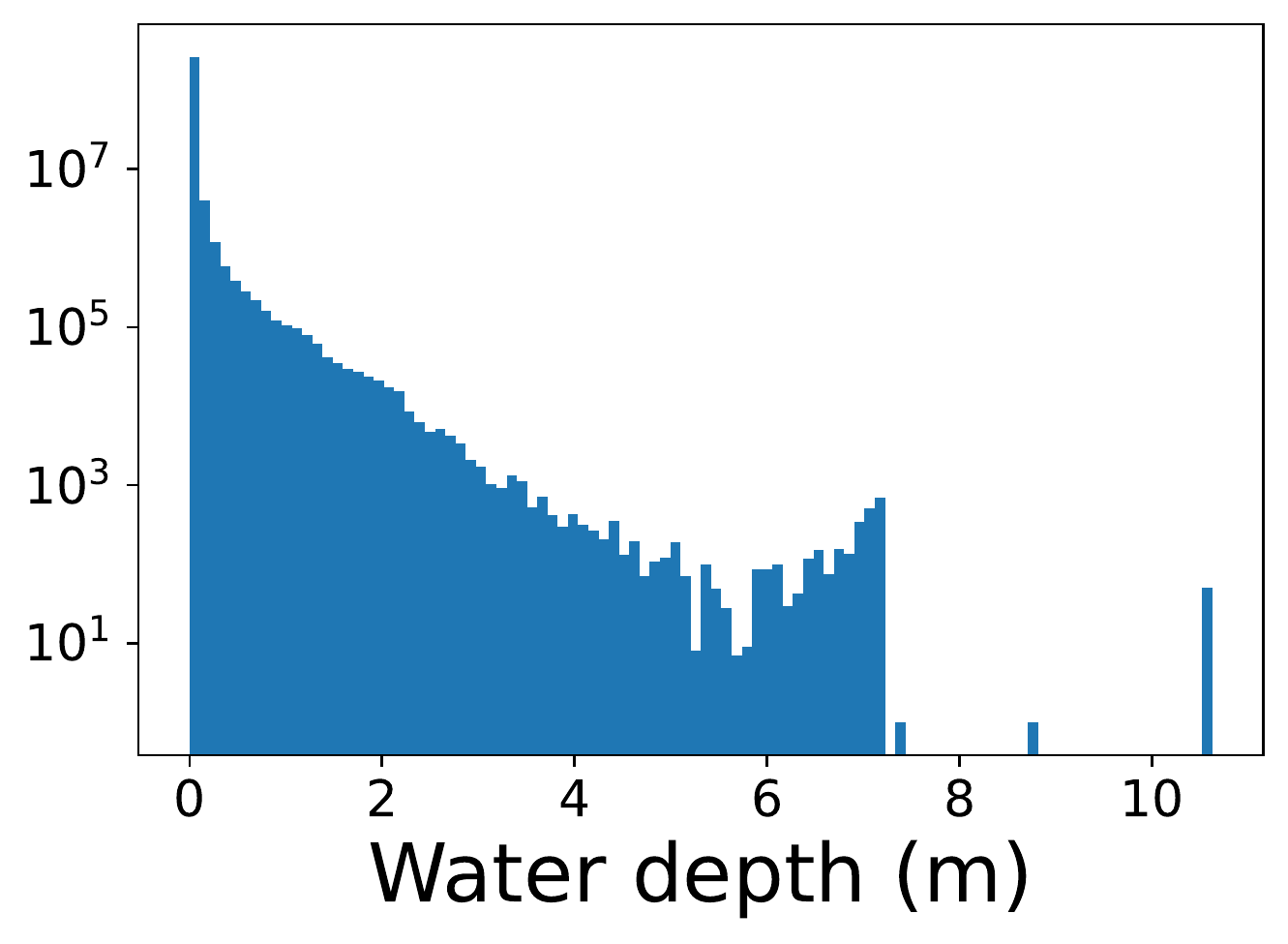}
    \caption{Histogram of water depths for event \textit{tr2\_1}.}
    \label{fig:hist}
\end{wrapfigure}

\begin{figure}[t]
    \centering
    \includegraphics[width=0.95\columnwidth]{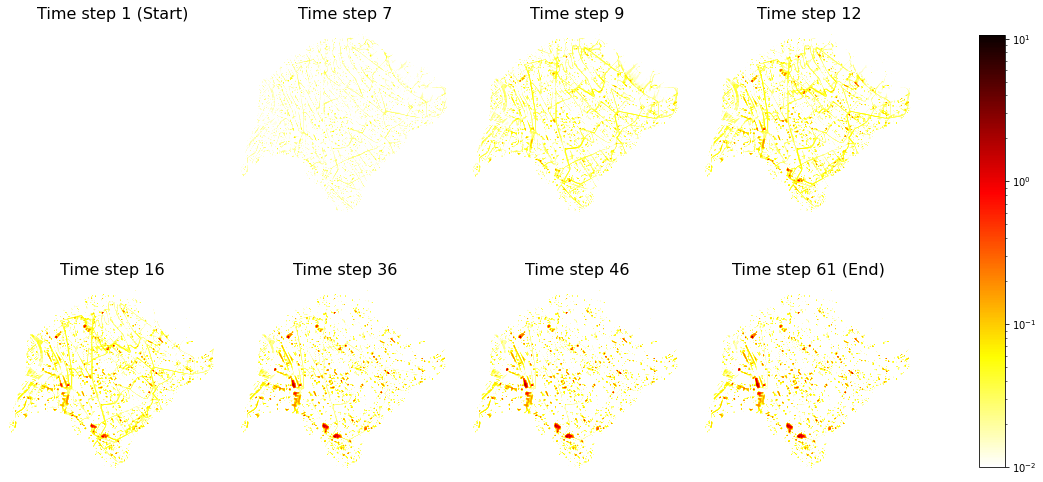}
    \caption[Evolution of spatial water depth over time]{Evolution of spatial water depth over time for event \textit{tr2\_1}.}
    \label{fig:spatialwd}
\end{figure}

The evolution of spatial water depths for one rainfall event is shown in Fig.~\ref{fig:spatialwd}. The time steps 1, 6, 8, 12, 16, 36, 46, and 61 are chosen to give an overview of the entire event. Note that one time step equals 5 minutes. It can be seen that the water depth quickly rises during time steps 1 to 12, and starts to decrease slowly after until the last time step 61. This is consistent with the rainfall event that only lasts for 60 minutes, which equals 12 time steps.

Note that most areas have very low water depths or no water at all (see Figure~\ref{fig:hist}). Most of the water will be densely concentrated in a few small areas. These areas are also the most critical ones, and therefore we care most about correctly predicting these extreme water depths.

In Section~\ref{sec:results}, we will also show one experiment on catchment area 744. However, because that data is structured and preprocessed in the same way as the one for catchment area 709, we will omit its details from the next paragraphs. 

\paragraph{Data Preprocessing}\label{sec:dpre}
Five features are selected and concatenated as input data $X$ for the models, including the DEM elevation $D$, spatial differential DEM elevation $\Delta D$, rainfall $R$, water depth $W$, as well as temporal differential water depth $\Delta W$ if the number of given input time steps is greater than 1.

These features are concatenated as multi-channel images (rasters), which are composed of \textit{3×T+H+3} channels, where \textit{T} is the number of input time steps ordered from oldest to latest, and \textit{H} is the amount of time steps we want to predict ahead for. The resolution of the catchment area 709 is \textit{2525×3000}, resulting in the corresponding shapes of the features, as shown in the Table \ref{tab:features}.

\begin{table}[h!]
\centering
\begin{tabular}{@{}lll@{}}
\toprule
\textbf{Variable} & \textbf{Feature} & \textbf{Shape}   \\ \midrule
$D$ & DEM        & 2525×3000×$1$      \\
$\Delta D$ & spatial differential DEM      & 2525×3000×$4$     \\
$R$ & rainfall    & 2525×3000×$(T+H-1)$     \\
$W$ & water depth        & 2525×3000×$T$     \\
$\Delta W$ & temporal differential water depth   & 2525×3000×$(T-1)$ \\
\bottomrule
\end{tabular}
\caption[Shape of the input features]{\label{tab:features}Shape of the input features for catchment area 709. The input $X$ consist of the concatenation of some or all of these features.}
\end{table}

The process to obtain $\Delta D$ and $\Delta W$ is as follows:
\begin{itemize}
    \item $\Delta D$: The DEM can be viewed as a 2D grid, where the elevation change can be measured in four directions: left, right, down and up. For each direction, we pad $D$ with -1 on the respective side before performing the subtractions along the columns ($c$) or rows ($r$) to introduce the correct shift in the output and to obtain the $\Delta D$ with the same resolution as~$D$:
    
    \begin{equation}
        \Delta D_1^{(i)} = c^{(i)} - c^{(i-1)} \text{  (Leftward change)}
    \end{equation}
    \begin{equation}
        \Delta D_2^{(i)} = c^{(i)} - c^{(i+1)} \text{  (Rightward change)}
    \end{equation}
    \begin{equation}
        \Delta D_3^{(j)} = r^{(j)} - r^{(j-1)} \text{  (Downward change)}
    \end{equation}
    \begin{equation}
        \Delta D_4^{(j)} = r^{(j)} - r^{(j + 1)} \text{  (Upward change)}
    \end{equation}
    where i and j are the column and row indices in the range [1, 3000] and [1, 2525], respectively. Remember that because of the padding, $c^{(0)}$, $r^{(0)}$, $c^{(3001)}$, and $r^{(2526)}$ denote added vectors which all equal $-1$.
    
    \item $\Delta W$: There are a series of spatial water depths for each rainfall event, and the water depths $W$ of adjacent time steps can be subtracted in time order. Then, $\Delta W$ with \textit{T-1} channels can be computed as follows:
    
    \begin{equation}\label{diffwd}
        \Delta W^{(t)} = W^{(t+1)} - W^{(t)},
    \end{equation}
    where t is the time step in the range [1, \textit{T}-1].
    
\end{itemize}

Note that the selection of these features was inspired by the work of~\cite{int_3}.
Naturally, we cannot handle the full input at once due to its size. Therefore, the input data $X$ is sampled in patches size of \textit{128×128}. In addition to reducing the input size, training on a large amount of randomly sampled patches can also help the models to generalize better.

\subsection{Neural network models}
\begin{figure}[ht]
    \centering
    \includegraphics[width=0.5\textwidth]{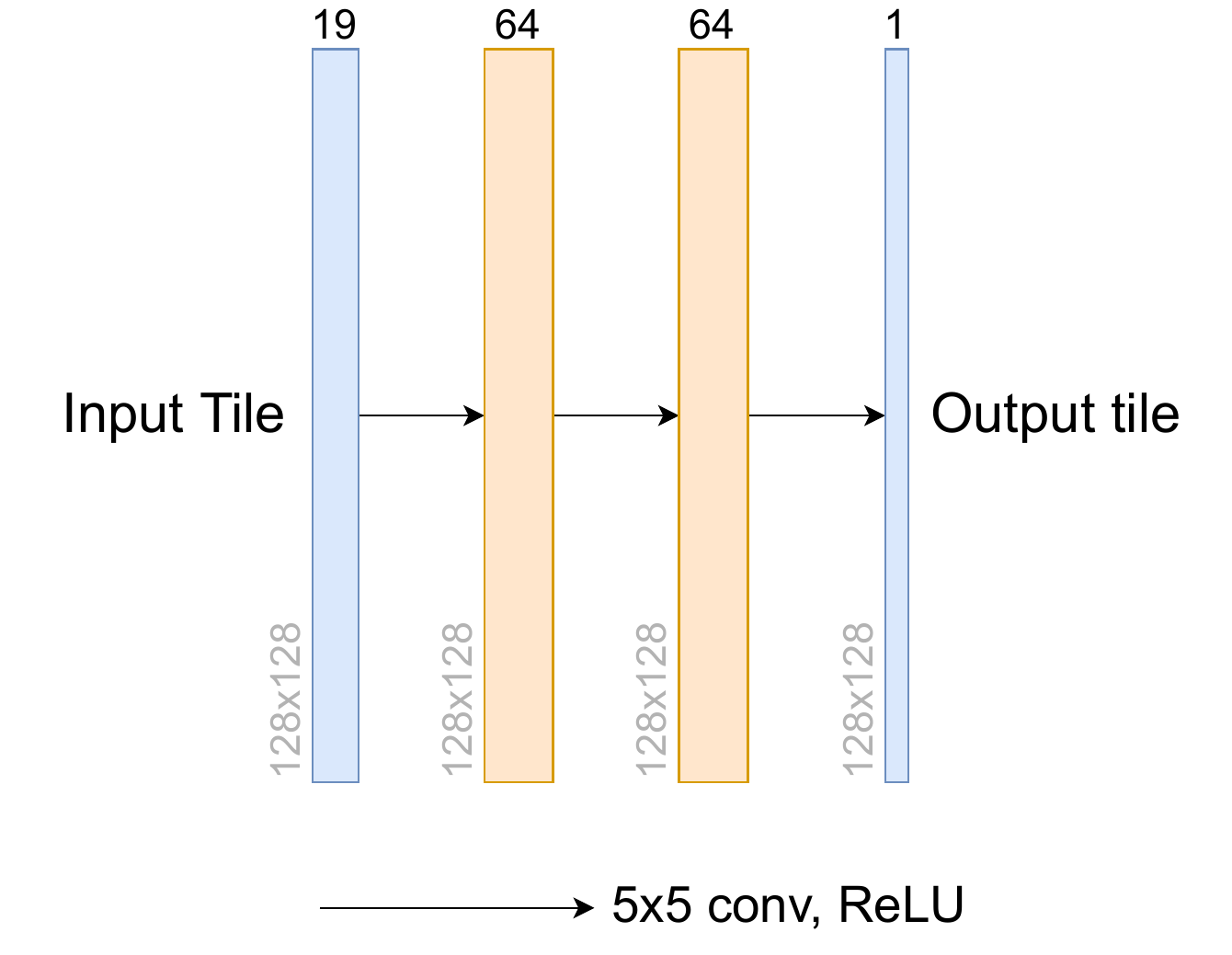}
    \caption{Overview of the simple FCN architecture. The spatial dimensions of the patch are preserved throughout the network.}
    \label{fig:FCN}
\end{figure}
In this subsection, we describe our neural network architectures. Each model is given the same inputs for a fixed amount of previous time steps $T$ and a fixed amount of future time steps H, which we want to predict ahead for (see Table~\ref{tab:features}). All data is provided for a catchment patch of size \textit{128x128} and concatenated. Finally, the models output a patch of equal size that predicts the amount of change in water depth for each pixel. Predicting changes instead of directly predicting water depths facilitates training. However, one can easily retrieve the actual water depths from these values.

\paragraph{Fully Convolutional Network (FCN)}
As a first, relatively simple model, we have chosen to implement a Fully Convolutional Network (FCN). The FCN consists of three convolutional layers with a kernel size of 5 and adequate zero padding on the sides to preserve the input dimensions (see Figure~\ref{fig:FCN}).

\paragraph{AutoEncoder}

\begin{figure}[ht]
    \centering
    \includegraphics[width=0.95\textwidth]{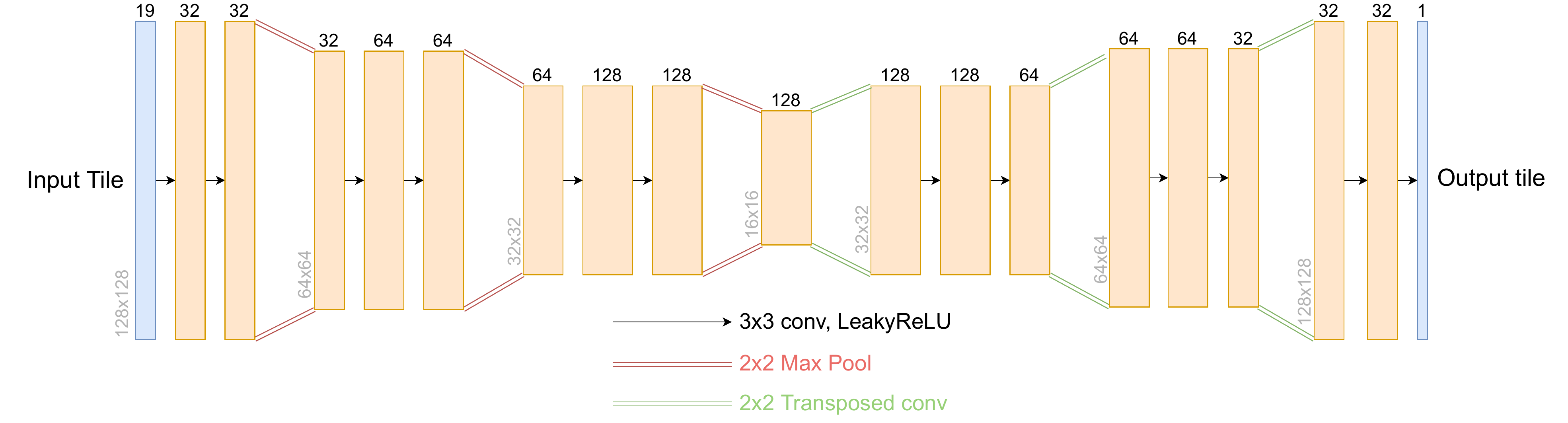}
    \caption{Overview of the CNN AutoEncoder architecture. The spatial dimensions of the full image with K input features (depending on amount of time steps given and other settings, and equaling 19 for our experiments) are first reduced through pooling operations. Later, they are restored through transposed convolutions to produce an output of the original input size.}
    \label{fig:Autoencoder}
\end{figure}

Our second network has been inspired by \cite{guo2021data}, who use a Convolutional AutoEncoder, which is another fully convolutional architecture. The network first increases the amount of features while decreasing the spatial dimensions through pooling operations. This part of the model is called the encoder as it ``encodes'' information into a reduced representation. In the second half of the network -- the decoder -- the spatial dimensions are restored through transposed convolutions. The architecture is visualized in Figure~\ref{fig:Autoencoder}. 

\paragraph{U-Net}

\begin{figure}[ht]
    \centering
    \includegraphics[width=0.8\textwidth]{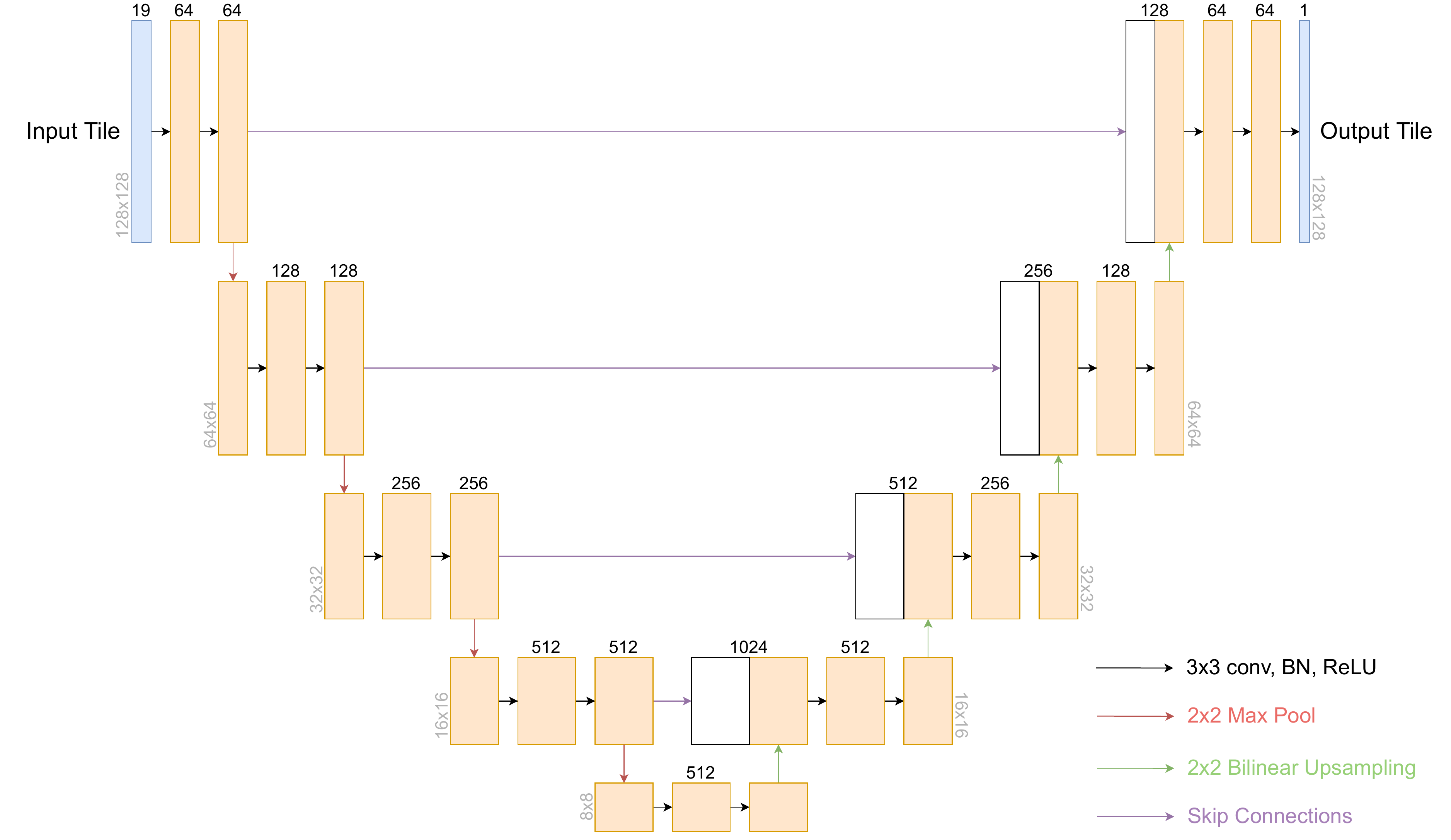}
    \caption{Overview over the U-Net architecture. The spatial dimensions of input are first downsampled through pooling operations and later upsampled through bi-linear upsampling. The skip connections seen in this figure give the architecture its characteristic U-shape.}
    \label{fig:U-Net}
\end{figure}

The U-Net architecture~\cite{ronneberger2015u} is similar to the CNN AutoEncoder in the sense that the spatial dimensions of the input are first down- and later upsampled to produce an output of the original dimensions. However, there are a few key changes to the architecture. Most importantly, the U-Net utilizes skip connections to concatenate the outputs of the left and right half of the network. Furthermore, we have decided to use bi-linear upsampling instead of transposed convolutions. The U-Net architecture is visualized in Figure~\ref{fig:U-Net}. 

\begin{figure}[ht]
    \centering
    \includegraphics[width=0.6\textwidth]{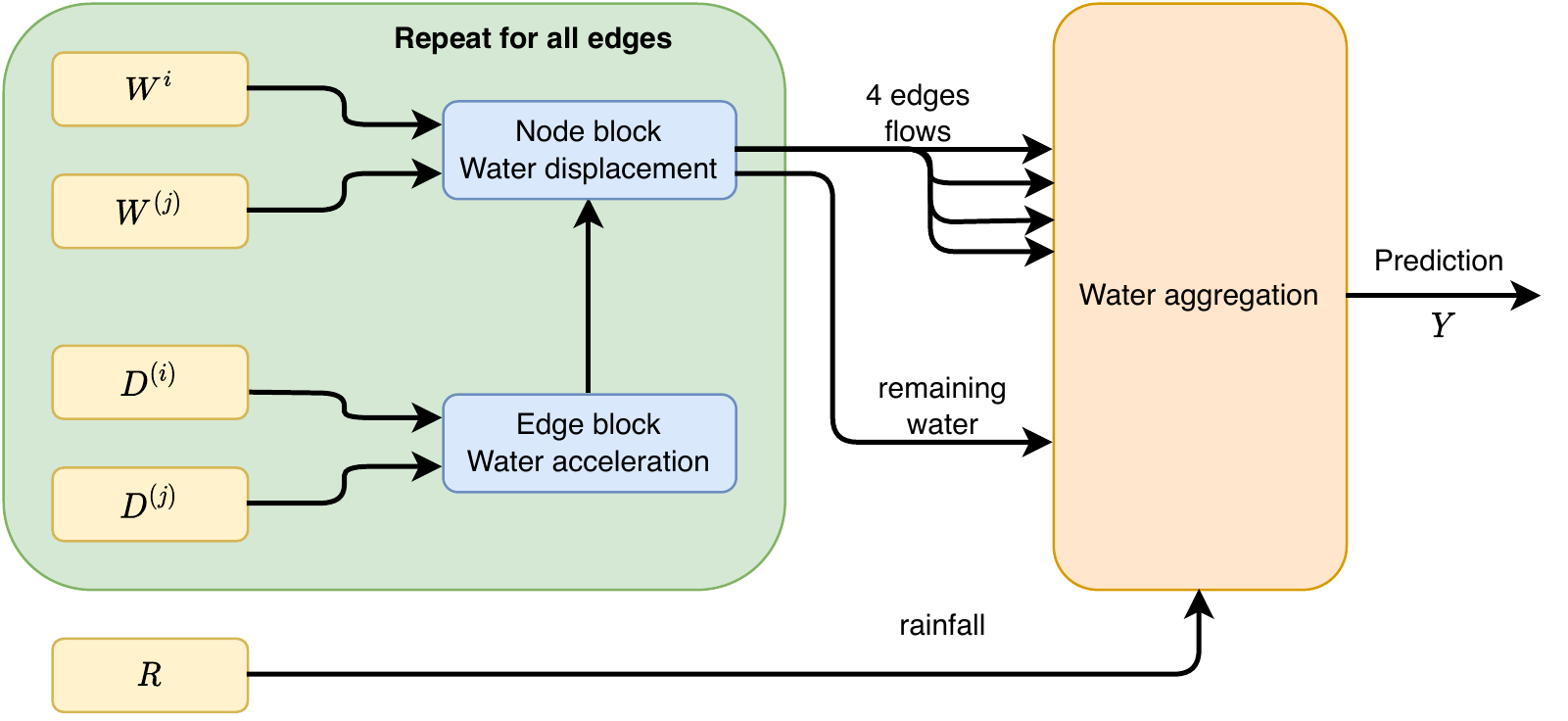}
    \caption{Simplified scheme of the graph architecture. The only learnable blocks are in blue. They can be seen as MLPs. In a first step, we compute the "water acceleration" based on the DEM in the edge block. Then, we combine this information with the water level on neighboring nodes in the "node block". Eventually, the output is one flow per edge and some remaining water on the node. Finally, we aggregate the water with a simple sum (orange block).}
    \label{fig:graph-model}
\end{figure}

\paragraph{Graph} Our graph model was designed with the specific goal of mimicking the underlying physical process of water flooding. To do so, we view the image as a graph where each pixel is a node with only 4 edges: up, down, left and right connection. This graph allows us to easily encode two simple hypotheses into the model: \textit{1)} the waterflow between two nodes depends only the elevation difference and the current amount of water on both nodes.\textit{ 2)} The resulting waterlevel on a node consists in the sum of the 4 waterflows (from the edges) and the remaining water on the node. A simplified\footnote{For simplicity, we omitted some of the input features in the network} version of the model is provided in Figure~\ref{fig:graph-model}. The learnable parameters of the network are the two Multilayer Linear Perceptrons (MLP) in blue. For efficiency reasons, we implemented these MLPs using 1x1 and 1x2 convolutions.

\subsection{Baselines}

To better evaluate the performance of our neural network models, we have implemented two non-parametric baselines. 
\paragraph{A. No change.} Our first baseline outputs no change in water depth at the next predicted time step. Assuming $T$ given time steps: 
\begin{equation}
    Y = W^{(t)},
\end{equation}
with $Y$ denoting the output (note that $T$ is the latest time step). Equivalently,
\begin{equation}
    \Delta Y = 0, 
\end{equation}
with $\Delta Y = Y - W^{(t)}$. 

\paragraph{B. Linear extrapolation.} The second baseline performs linear extrapolation, i.e.\ it predicts
\begin{equation}
    Y = W^{(t)} + \Delta W^{(t-1)},
\end{equation}
or, equivalently, 
\begin{equation}
    \Delta Y^{(t)} = \Delta W^{(t-1)}.
\end{equation}
Note that $\Delta W^{(t-1)} = W^{(t)} - W^{(t-1)}$.


Furthermore, we also implemented two baselines using classic autoregressive models. These models are only given the water depth data without any information on the DEM or the rainfall data. The incentive for leaving out this information here is to test whether the more complex models are able to sufficiently utilize this additional data for improved predictions over these baselines.

\paragraph{C. Autoregressive (AR) 1x1.} We implement a single convolutional layer with a 1x1 kernel to predict the next water depths in a pixel-wise manner. 
\paragraph{D. Autoregressive (AR) 5x5.} As a second model, we implement a single convolutional layer but with a 5x5 kernel to take into account some information of close-by locations. 

The models are only trained to predict the water depths in the next time step. If we require a prediction for a later time step, we apply the model autoregressively until we arrive at this point in time. To be precise, each output of the model will be added to the input in place for the oldest time step in a looping manner until we reach the desired time step. 

\subsection{Loss function} \label{sec:losses}
The loss function, which is optimized during training, measures the quality of a set of network parameters w.r.t. the training data.
For this loss function, we have experimented with several variations of the mean absolute error (MAE) and mean squared error (MSE). Finally, due to difficulties with learning the largest differences in water depths (which are considered the most important predictions), we have settled for a variant of MAE that increases the loss by a factor of 4 wherever the target water depth changes by more than~20cm. As it can be observed in Figure~\ref{fig:hist}, the amount of large water depths is relatively small but most critical for our case. However, in practice such water levels are of the highest importance since they correspond to the most dangerous situations.

\subsection{Evaluation method}
\label{sec:evaluation_method}
We predict water depths for the test rainfall events that have not been seen during training to test the performance of the implemented models. Since our main interest is to avoid errors at higher water depths, we define different water depth ranges: our main evaluation metric $M$ measures the absolute errors within 5 buckets (intervals) depending on the ground truths, for water depths of 0-10cm, 10-20cm, 20-50cm, 50-100cm, and $>$100cm, respectively. Since the errors will vary greatly between buckets, we further normalize them by the standard deviation of the ground truths within each bucket. To summarize, we compute for every pixel-wise prediction $\hat{y}_i$ in each bucket~$b$:
\begin{equation} \label{eq:metric}
    M_b(y_i,\hat{y}_i) = \frac{|y_i - \hat{y}_i|}{\sigma_b},
\end{equation}
where $y$ is a vector of all ground truths in bucket $b$, and $\sigma_b$ denotes the standard deviation over these ground truths. Naturally, we want to minimize this metric.

The normalization (division by the standard ground truth deviation) used in our metric, i.e. \eqref{eq:metric} has the advantage to provide an error theoretically scale-independent of the water level. Furthermore, similar to the coefficient of determination, this metric quantifies globally learning in comparison with a mean predictor. The error of a mean predictor is in average 1 as the numerator will be, in average, the standard deviation. Therefore, we consider that the models are learning if the metric \eqref{eq:metric}  is below 1 and that the model is failing to learn if the metric is above 1.

\section{Results} \label{sec:results}

Through testing many different configurations, we have found the following settings that work well across all of the models. Each model receives 5 previous time steps as input, where the spatial dimension of the features is of size 128x128, and predicts the change in water depth one hour ahead (12 time steps of 5 minutes). All parametric models are trained for 30 epochs with a learning rate of $10^{-3}$ and the Adam optimizer \cite{kingma2014adam}, except the AutoEncoder, which we train for 50 epochs with a learning rate of $10^{-4}$ because it failed to train with a higher learning rate.

\subsection{Training methods}

To facilitate further evaluations on all networks and baselines, we first aim to establish a preferred training method. We consider three options:
\begin{enumerate}[label=\Alph*.]
    \item \textbf{1 TS}: The network is trained to predict one time step ahead and evaluated iteratively to get predictions for 12 time steps ahead.
    \item \textbf{12 TS Direct}: The network is trained to directly predict 12 time steps ahead.
    \item \textbf{12 TS Iterative}: The network is trained to predict 12 time steps ahead iteratively, i.e.\ already during training it uses predictions for one time step as an additional input for predicting the next time step until all 12 time steps are predicted.
\end{enumerate}

\begin{figure}[ht]
    \centering
    \begin{subfigure}[ht]{0.48\textwidth}
        \includegraphics[width=\textwidth]{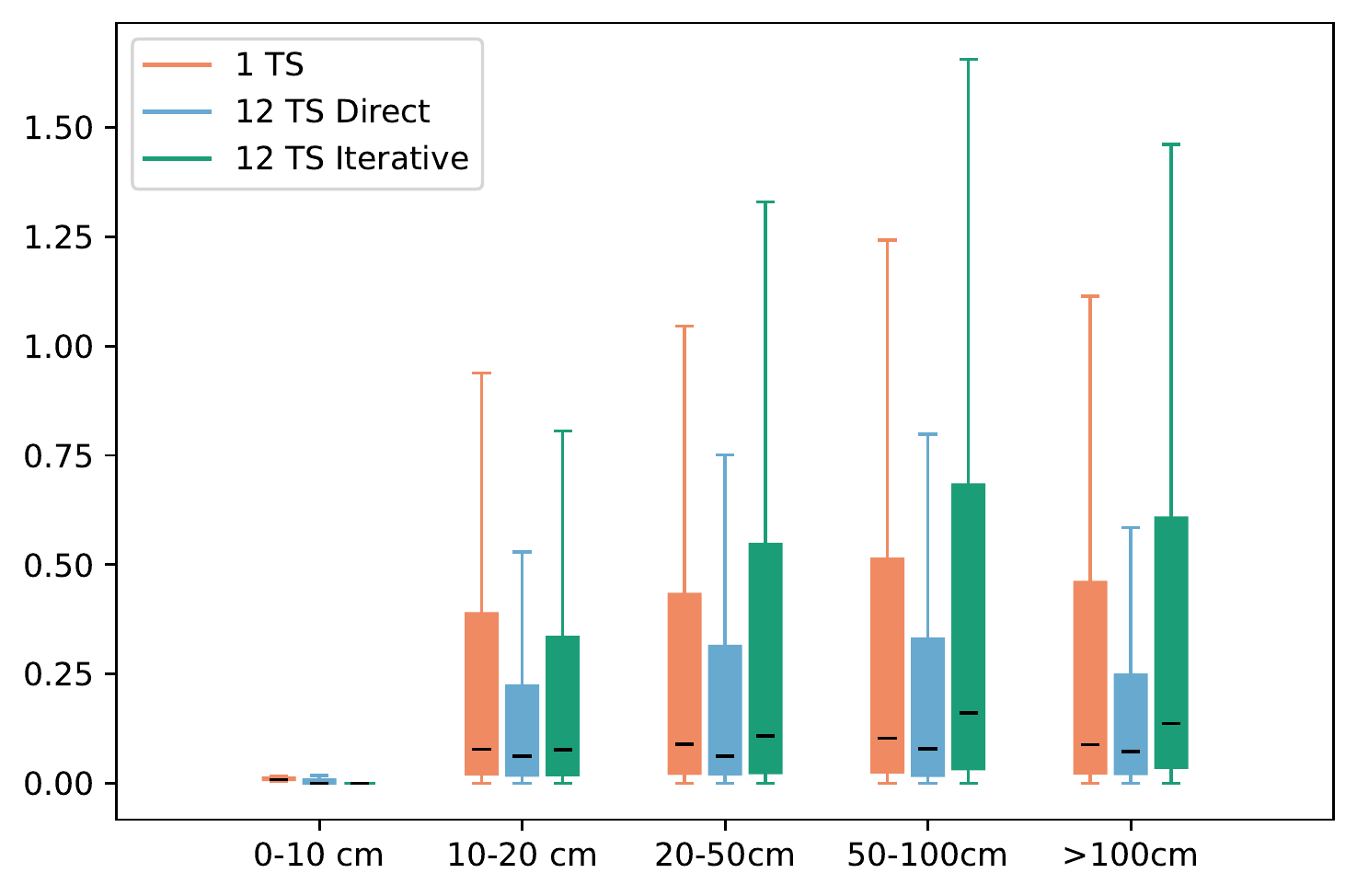}
    \end{subfigure}
    \hfill
    \begin{subfigure}[ht]{0.48\textwidth}
        \includegraphics[width=\textwidth]{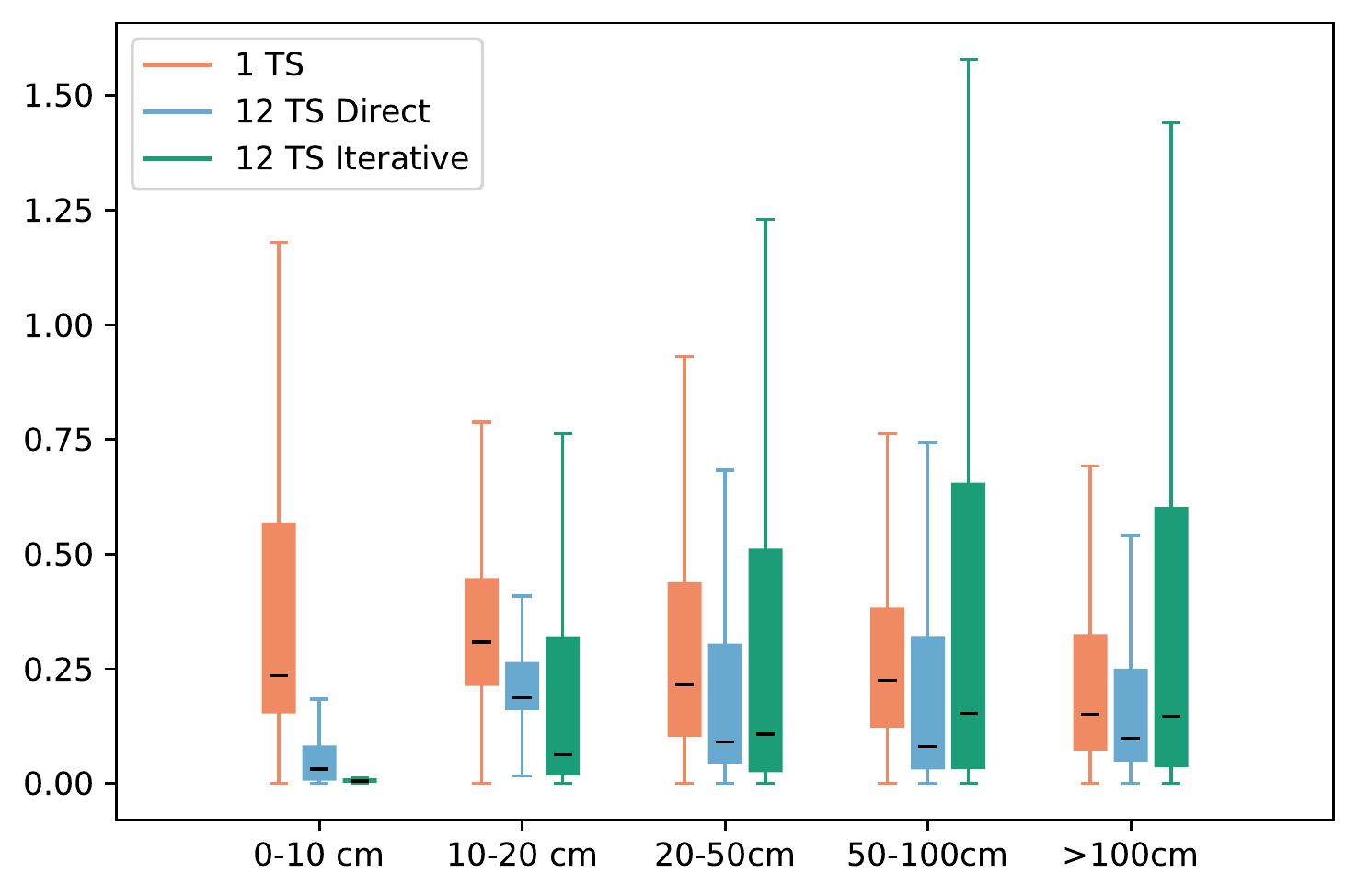}
    \end{subfigure}
    \caption{Comparison of training methods for the FCN (left) and Graph (right) models. The error metric is described in Section~\ref{sec:evaluation_method}. Both models have been trained and evaluated on the 709 short rainfall events (see Figure~\ref{fig:hyetographs}).}
    \label{fig:training_methods}
\end{figure}

We have implemented all training methods for two of our models with the hope that the results will be sufficiently predicative to assume that the behavior will generalize to the other models as well. To this end, we have evaluated the training methods for our FCN as well as the Graph model in Figure~\ref{fig:training_methods}. Despite some issues with low water depths, we can conclude that training to directly predict 12 time steps ahead yields clearly better results than the other training methods, especially for larger water depths, which are the most critical ones. This comes with the additional benefit that direct prediction is significantly faster than autoregressively predicting one step after another. Therefore, all models in subsequent experiments will use this training method, except the last two baselines that are designed to be applied autoregressively.

\subsection{Model comparison}

In this section, we evaluate all four deep models, the two non-parametric baselines, as well as our two versions of the autoregressive baseline. Out of all experiments that have been performed, we chose three that should give the most complete picture over the overall performance of the models.

\paragraph{Short rainfall events}

For our first evaluation, we compare the  models' performance on the short events of the 709 catchment. The specific training, validation and testing split that we use for this experiment can be seen in Table~\ref{tab:dataset}.

\begin{figure}[ht!]
    \centering
    \includegraphics[width=\textwidth]{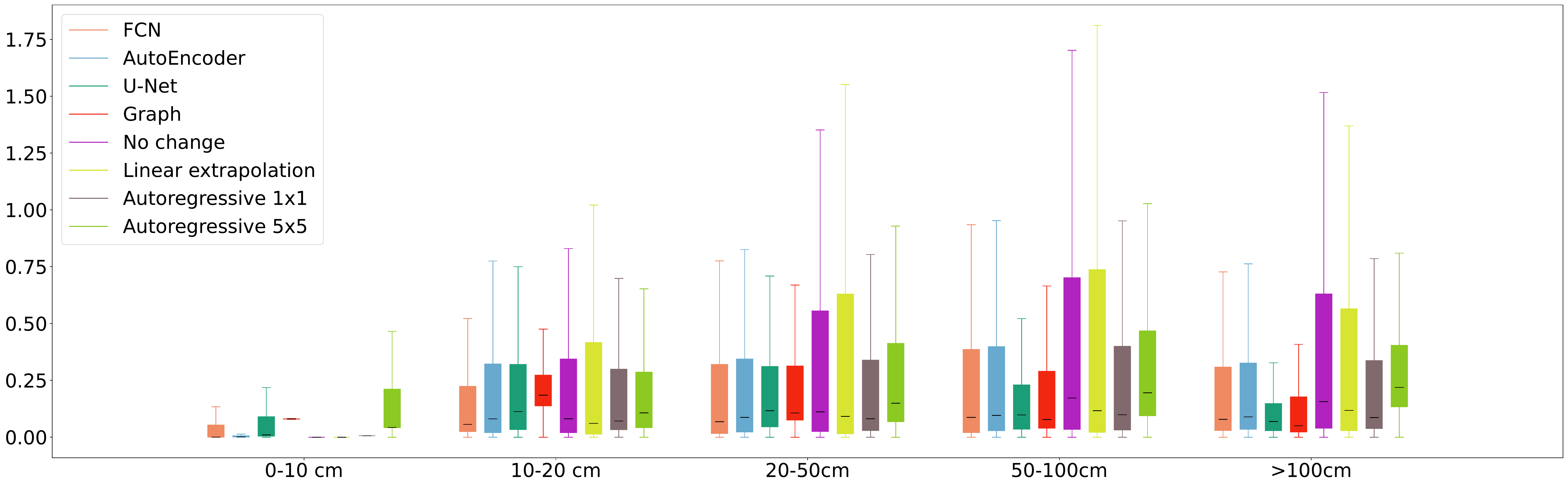}
    \caption{Box plots for all models trained and tested on catchment area 709 with short events only. The medians are denoted by a black dash.}
    \label{fig:model-comparison-short}
\end{figure}

Our results are summarized in Figure~\ref{fig:model-comparison-short}. For the lower water depths, some of the baseline models outperform the deep networks. Under the assumption that lower water depths also show less change over time, it is not too surprising that predicting no change at all might result in relatively small errors overall. For the two non-parametric baselines, this behavior is certain, but also the simple autoregressive baselines seem to be able to capture little to no change better than the more involved models. However, we care about more about predictions for large water depths which are more critical. In the range of 50-100cm and especially for water depths of more than 1m, we see that both the U-Net as well as the graph model show significantly lower errors compared to the other methods.

\begin{table}[ht]
\centering
\begin{tabular}{@{}cc@{}}
\toprule
\textbf{Dataset} & \textbf{Rainfall Events} \\ \midrule
Training Set     & \begin{tabular}[c]{@{}c@{}}tr5\_1, tr20\_1, tr50\_1, tr2\_2, tr10\_2, \\tr20\_2, tr50\_2, tr5\_3, tr10\_3, tr100\_3\end{tabular} \\ \midrule
Validation Set   & tr100\_2, tr2\_3         \\ \midrule
Test Set         & tr2\_1, tr10\_1, tr100\_1, tr5\_2, tr20\_3, tr50\_3  \\ \bottomrule
\end{tabular}
\captionsetup{width=0.7\textwidth}
\caption{Allocation of training, validation and test set for the 709 catchment with only short rainfall events.}
\label{tab:dataset}
\end{table}

\paragraph{Long rainfall events}

In a next step, we want to introduce long rainfall events to the training and testing procedures. We add the events \textit{real1\_c1} and \textit{tr50\_3c2} (see rightmost plot in Figure~\ref{fig:hyetographs}) to the training set from Table~\ref{tab:dataset} and event \textit{tr50\_3c1} to the validation set. As the test set, we replace the two short events with the long event \textit{real2\_c1}. We decided to replace the short events in the test set because we want to measure the performance on a long event specifically, without influence from the possibly different performance on the short events.

\begin{figure}[ht!]
    \centering
    \includegraphics[width=\textwidth]{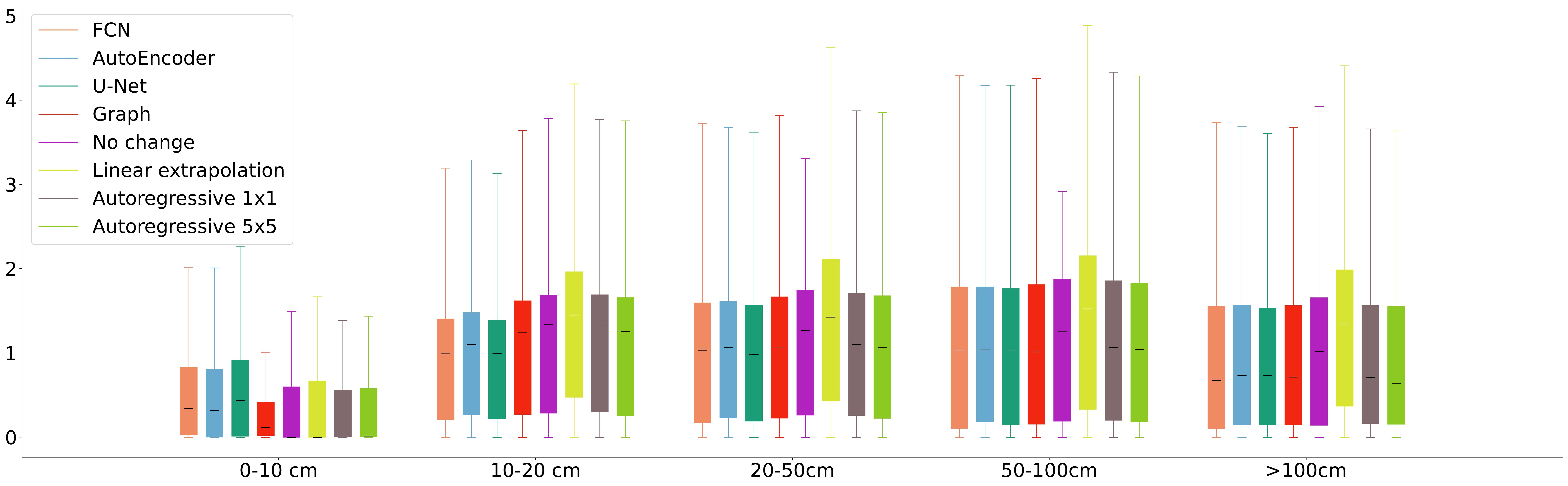}
    \caption{Box plots for all models trained and tested on catchment area 709 with both short and long events. The medians are denoted by a black dash.}
    \label{fig:model-comparison-long}
\end{figure}

From the results from Figure~\ref{fig:model-comparison-long} we see that the errors are considerably higher for all models compared to the short events. Unfortunately, we cannot observe any advantages from the more involved models over a simple model like the FCN or even over the autoregressive baselines. There is a small, but -- due to the overall poor performance -- insignificant benefit over the non-parametric baselines that predict no change at all or linearly interpolate the water depth.

\paragraph{Generalization to different catchment}

Finally, we want to evaluate the models' capabilities of generalizing to a different catchment area. To this end, we take the same trained models from the last paragraph (i.e.\ trained on catchment area 709 with both short and long events) and test them on a long event for catchment area 744. The event chosen has the same rainfall pattern as the long event \textit{real1\_c1} from catchment area 709.

\begin{figure}[ht!]
    \centering
    \includegraphics[width=\textwidth]{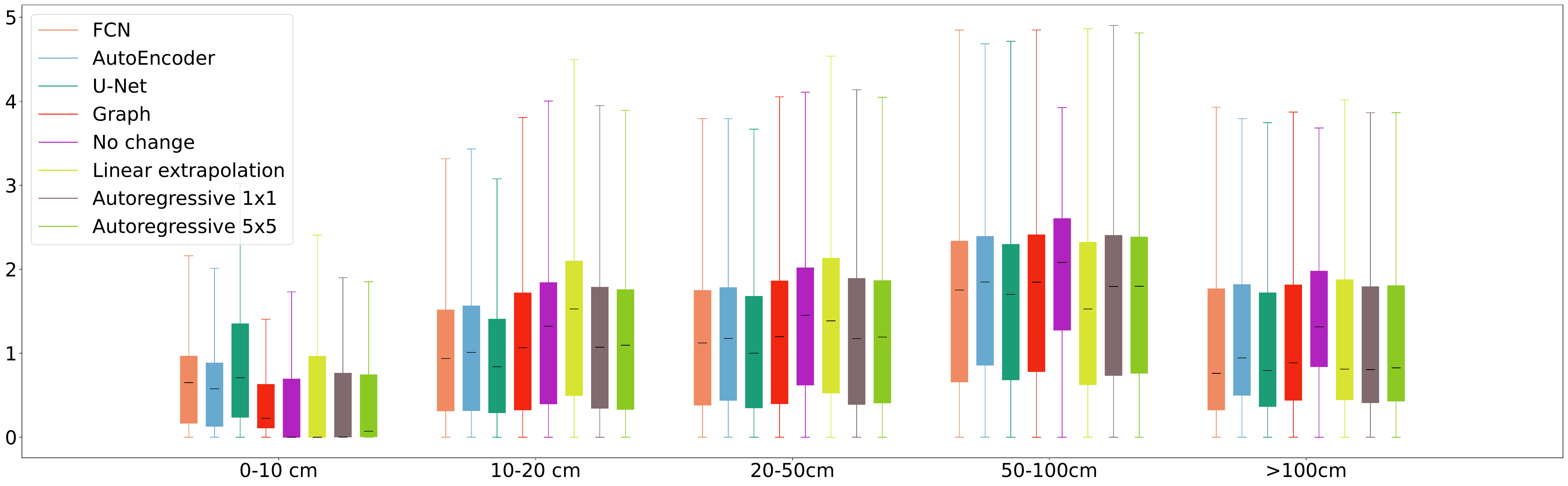}
    \caption{Box plots for all models trained on catchment area 709 with both short and long events and tested on one long event from catchment 744. The medians are denoted by a black dash.}
    \label{fig:model-comparison-744}
\end{figure}

The results for this experiment are presented in Figure~\ref{fig:model-comparison-744}. Once again, the deep models do not show any benefits over the simpler ones. For the largest water depths, even linear extrapolation of the water depths is on par with the predictions from parameterized models.

We have performed further tests on other long events, and experimented with training on long events only. Unfortunately, those additional experiments have also not yielded dramatically changed results. This leaves the more involved, deep models with improved results only on the short rainfall events but without any advantages on the experiments with long events for either catchment area.

\section{Discussion}
\label{sec:discussion}
In Figure \ref{fig:model-comparison-short}, where only short events are used in the training set, the metric values are significantly below 1 indicating that the models are learning the distribution and generalizing to unseen data. We also observe that learned models are better than the baselines, especially for high water levels ($>0.5m$). U-Net is overall the best model. While the graph network is the best for high water levels, it has bad performance for low water levels. Despite the fact that the reconstruction error is low, we are not satisfied with the proposed networks as they are not significantly better than the baselines. This probably indicates that the networks only capture the easiest correlations in the dataset.

In Figure \ref{fig:model-comparison-long} and Figure \ref{fig:model-comparison-744}, we evaluate the model on two different test sets with distribution shifts. In Figure \ref{fig:model-comparison-long}, we observe that for long rainfall events the performance drops significantly (the metric is above 1), meaning that the model does not generalize.
Similarly, in the case of a different catchment (Figure \ref{fig:model-comparison-744}), the model is also not able to generalize. Our interpretation is that long rainfall events likely include more complex dynamics that the networks were not able to learn. This hypothesis is supported by the fact that the performance of non-learnable baselines has also dropped significantly compared to the short rainfall event test set. One final observation is that on these more challenging test sets, the networks are not improving over the baselines.

\section{Conclusion}
As discussed in the previous Section~\ref{sec:discussion}, our current model provides unsatisfying results for two reasons. First the overall performance improvement of the proposed models compared to the baselines is insufficient to justify their additional complexity. Second, the proposed models are not able to generalize to events outside of the training distribution, making them unsuitable for a real use case. 

One possible problem with the current setup might be the lack of diversity in the training rainfall events. While the models do not overfit the data per se (i.e. the loss does not drop below a constant significantly above 0), it is likely that the models overfit some characteristics of the training distribution, preventing them from generalizing to different rainfall events. Unfortunately, generating a training set with a high rainfall diversity would be very computationally expensive.
Another potential weakness of our setup lies in the fact that the proposed model are more appropriate to exploit space correlations instead of time correlations. This choice might not be suitable for modeling water flows, which are affected by complex time dynamics. Therefore, the problem could benefit from architectures more designed for time series, such as recurrent networks~\cite{hermans2013training}, dilated convolutions~\cite{dutilleux1990implementation}, or transformers~\cite{vaswani2017attention}. 

Possibly, the physics of flood simulation might be too challenging to learn from scratch and a different approach might be preferable. They are multiple "physically inspired" contributions~\cite{frame2022deep, beucler2021enforcing, ruckstuhl2021training, cheng2021deep} leveraging using the structure of a simulator with some parts being replaced with a neural network. Typically, this would allow to work with large time step (in comparison with the original simulator) and significantly accelerate the simulation process. 
In the interest of reproducibility, we make our code and datasets available at  {\color{red}\url{} and \url{}}.

\newpage

\bibliographystyle{abbrv}
\bibliography{4real.bib}

\end{document}